\documentclass[10pt,twocolumn,letterpaper]{article}

\usepackage[pagenumbers]{cvpr} 

\definecolor{drp-blue}{HTML}{1f77b4}
\definecolor{pretty-blue}{RGB}{0, 113, 188}
\definecolor{kaiming-green}{RGB}{57,181,74} 
\definecolor{mypurple}{RGB}{55,0,168} 
\definecolor{icmlblue}{rgb}{0,0.08,0.45} 
\definecolor{linecolor1}{HTML}{F1F7FB}
\definecolor{linecolor2}{HTML}{E3EFF7}
\definecolor{linecolor3}{HTML}{D5E4F0}

\definecolor{reconcolor}{HTML}{412F8A}
\definecolor{runpei-orange}{HTML}{F35F27}
\definecolor{runpei_blue}{HTML}{14294B}
\definecolor{datacolor}{HTML}{0009BF}
\definecolor{vitcolor}{HTML}{fc8e62}
\definecolor{xycolor}{HTML}{EF98AA}
\definecolor{lightpink}{HTML}{F9F0EE}
\definecolor{deeppink}{HTML}{EBD9E4}
\definecolor{lightpurple}{RGB}{123,107,143}
\definecolor{lightyellow}{RGB}{200,180,120}
\definecolor{MorandiLightBlue}{RGB}{150,180,220}
\definecolor{MorandiLighterBlue}{RGB}{230,240,250}
\definecolor{MorandiPink}{RGB}{200, 160, 180}
\definecolor{MorandiLightPink}{RGB}{250, 240, 245}
\usepackage{pifont}
\def\baselineexp{{\scshape Base}\xspace}
\def\Noexp{{\scshape No Think}\xspace}
\def\codexp{{\scshape CoD}\xspace}
\def\alphaexp{{\scshape Alpha-1}\xspace}
\def\chainvexp{{\scshape ChainV}\xspace}

\definecolor{numb}{rgb}{0.2,0.2,0.7}
\definecolor{punct}{rgb}{0.7,0.2,0.2}
\definecolor{delim}{rgb}{0,0.5,0}
\definecolor{background}{rgb}{0.95,0.95,0.95}

\usepackage{listings}
\usepackage{xcolor}

\lstdefinelanguage{json}{
    basicstyle=\ttfamily\small,
    showstringspaces=false,
    breaklines=true,
    frame=single,
    backgroundcolor=\color{gray!5},
    literate=
     *{0}{{{\color{numb}0}}}{1}
      {1}{{{\color{numb}1}}}{1}
      {2}{{{\color{numb}2}}}{1}
      {3}{{{\color{numb}3}}}{1}
      {4}{{{\color{numb}4}}}{1}
      {5}{{{\color{numb}5}}}{1}
      {6}{{{\color{numb}6}}}{1}
      {7}{{{\color{numb}7}}}{1}
      {8}{{{\color{numb}8}}}{1}
      {9}{{{\color{numb}9}}}{1}
      {:}{{{\color{punct}{:}}}}{1}
      {,}{{{\color{punct}{,}}}}{1}
      {\{}{{{\color{delim}{\{}}}}{1}
      {\}}{{{\color{delim}{\}}}}}{1}
      {[}{{{\color{delim}{[}}}}{1}
      {]}{{{\color{delim}{]}}}}{1},
}

\lstdefinestyle{jsonstyle}{
    language=json,
    basicstyle=\ttfamily\footnotesize,
    frame=single,
    backgroundcolor=\color{gray!5},
    breaklines=true
}

\definecolor{cvprblue}{rgb}{0.21,0.49,0.74}
\usepackage{multirow}
\usepackage{makecell}
\usepackage[table,xcdraw]{xcolor}
\usepackage[pagebackref,breaklinks,colorlinks,allcolors=cvprblue]{hyperref}

\usepackage{amsmath,amsfonts,bm}

\def\eqref#1{equation~\ref{#1}}

\def\1{\bm{1}}

\def\vx{{\bm{x}}}

\def\mA{{\bm{A}}}

\def\mH{{\bm{H}}}

\def\mM{{\bm{M}}}

\def\mT{{\bm{T}}}

\def\mW{{\bm{W}}}

\def\mZ{{\bm{Z}}}

\DeclareMathAlphabet{\mathsfit}{\encodingdefault}{\sfdefault}{m}{sl}
\SetMathAlphabet{\mathsfit}{bold}{\encodingdefault}{\sfdefault}{bx}{n}

\renewcommand{\a}{{\bm{a}}}

\renewcommand{\u}{{\bm{u}}}
\renewcommand{\v}{{\bm{v}}}

\def\sI{{\mathbb{I}}}

\def\sP{{\mathbb{P}}}

\def\sT{{\mathbb{T}}}

\usepackage{amsmath}

\def\paperID{2749} 
\def\confName{CVPR}
\def\confYear{2026}

\title{\chainvexp: Atomic Visual Hints Make Multimodal Reasoning Shorter and Better}

\author{Yuan Zhang${}^{1,2}$, Ming Lu${}^{1}$, Junwen Pan${}^{2}$ \\ Tao Huang${}^{3}$, Kuan Cheng${}^{1}$, Qi She${}^{2}$, Shanghang Zhang${}^{1\dag}$ \\
{\normalsize \textsuperscript{1}School of Computer Science,  Peking University \quad \textsuperscript{2}ByteDance Inc. \quad \textsuperscript{3}Shanghai Jiao Tong University}
}

\begin{document}
\maketitle
\addtocontents{toc}{\protect\setcounter{tocdepth}{0}}
\begin{abstract}
Recent advances in multimodal reasoning models have demonstrated impressive capabilities across text and vision. However, even leading models exhibit redundant self-reflection when generating lengthy reasoning chains. While training-free CoT compression methods have emerged in the LLMs domain, they rely on static visual references and thus provide limited gains for multimodal reasoning. Therefore, we propose ChainV, a framework that dynamically integrates visual hints into the reasoning process, thereby making multimodal reasoning shorter and better. Specifically, ChainV first performs a coarse visual patch selection based on the previous reasoning step, then refines it by identifying the most representative atomic visual hint according to the averaged attention intensity. Additionally, ChainV introduces a consistency-based evaluation mechanism to assess the reliability of the chosen hint, guiding the model to adaptively adjust its level of self-reflection. Eventually, the pixel coordinates of the selected visual hint and its reliability are incorporated into thinking with a Bernoulli stochastic process. Experiments indicate that our method significantly improves reasoning accuracy and efficiency, especially on math-intensive benchmarks where visual hints are crucial for multi-step symbolic reasoning. For example, ChainV achieves $2.3\%$ improvement on the MathVista within MIMO-VL-RL, while reducing inference latency by $51.4\%$ and shortening output token length by $24.5\%$.

\end{abstract}

\section{Introduction}
\label{sec:intro}




Chain-of-Thought (CoT) \cite{kojima2022large, lyu2023faithful, wang2024chain} has proven to be an effective mechanism for strengthening the reasoning ability of Large Language Models (LLMs) ~\cite{brown2020language, achiam2023gpt, touvron2023llama, peng2023instruction, bi2024deepseek} by encouraging them to generate explicit, step-by-step rationales. With the rise of Multimodal Large Language Models (MLLMs) \cite{wang2024qwen2, chen2024internvl, zhang2024unveiling, wu2024deepseek, zhang2024sparsevlm}, CoT has also become increasingly essential across a wide range of multimodal tasks \cite{xiaomi2025mimo, team2025kwai, yang2025qwen3, xiao2024logicvista, he2024olympiadbench}. It enables MLLMs to perform reasoning over textual and visual inputs, bridging low-level perceptual understanding with high-level abstract reasoning.

\begin{figure}
    \begin{center}
        \includegraphics[width=1.05\linewidth]{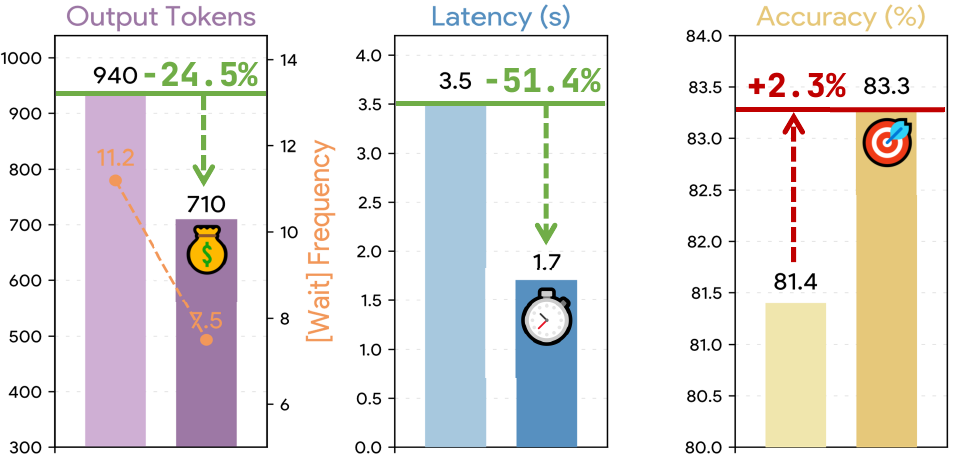}
    \end{center}
    \vspace{-4mm}
    \caption{\textbf{ChainV enables efficient multimodal reasoning.}  
    Results demonstrate shorter reasoning chains, lower inference latency, and improved accuracy with ChainV.}
    \label{fig:com}
    \vspace{-4mm}
\end{figure}

However, these performance gains are accompanied by a \textit{non-trivial efficiency cost}, as the autoregressive generation of lengthy reasoning tokens introduces considerable latency and memory overhead at inference time. For instance, in Figure~\ref{fig:com}, MiMo-VL-RL \cite{xiaomi2025mimo} requires an average of $3.5$s to respond to a single MathVista \cite{lu2023mathvista} question, consuming nearly $1,000$ tokens per answer. In deployment scenarios where user experience and response time are critical, such overhead becomes a practical bottleneck, rendering blind expansion of reasoning both inefficient and unsustainable.

To our knowledge, existing CoT efficiency techniques, whether \textit{training-free} or \textit{training-based}, have been developed almost exclusively for LLMs, leaving the multimodal setting largely unexplored.
Training-free methods \cite{xu2025chain, zhang2025alphaone, li2025thinkless}, \eg, reducing output length via prompt design \cite{xu2025chain} or manually inserting \texttt{<wait>} tokens \cite{zhang2025alphaone}, can be applied to MLLMs. However, these approaches perform chain-of-thought reasoning solely in language, relying on the same static visual references, and thus offer limited benefits for multimodal reasoning. In contrast, human reasoning naturally integrates both language and vision. Therefore, visual references should also dynamically evolve to align with the reasoning process.
Training-based methods \cite{wu2025arm, wang2025adareasoner, huang2025adacot}, on the other hand, typically rely on additional training, curated datasets, or supervised fine-tuning, thereby introducing substantial engineering complexity and resource overhead. Furthermore, their reliance on task-specific data or model customization limits generalizability, hindering scalability and deployment in real-world systems.

This reveals a clear gap: \textit{an efficient, training-free CoT mechanism that is explicitly grounded in visual evidence is still missing for multimodal reasoning}. In this work, we address this gap by introducing \textbf{ChainV}, a plug-and-play reasoning pipeline that leverages visual hints to adaptively regulate thinking depth, reducing unnecessary token generation without any retraining or architectural changes.

We begin by examining how multimodal reasoning models actually behave during deliberate “\textit{slow thinking}”.
Our analysis reveals that the model frequently emits a large number of \texttt{<wait>} tokens used for internal self-reflection, \eg, MiMo-VL-RL produces an average of $11.2$ such pauses on MathVista, without introducing any new visual evidence.
This indicates that the model is not lacking reasoning time, but rather lacking grounded visual guidance during reasoning.
Motivated by this observation, ChainV intervenes precisely at the “\textit{wait stage}”: instead of allowing the model to enter another redundant reflection loop, we probabilistically replace it with a visual hint. Concretely, ChainV first performs a coarse visual assistant patch selection based on the previous reasoning step, then refines it by selecting the most representative atomic hint according to the averaged answer attention intensity.
In addition, ChainV introduces a consistency-based evaluation mechanism to estimate the \textit{reliability} of the selected hint, to measure how faithfully it supports the thinking process. 
Eventually, the pixel-level coordinates of the selected atomic hint and its reliability are injected into the reasoning stream, enabling the model to recall grounded evidence.

ChainV leverages customized visual hints to reduce redundant reflection loops (“\textit{wait cycles}”), thereby shortening reasoning traces, improving accuracy, and lowering token and time costs compared to full CoT decoding (Figure~\ref{fig:com}). Concretely, when integrated with Keye-VL-Thinking 8B \cite{team2025kwai}, ChainV reduces inference latency by \textbf{44.7\%} on $\text{MMMU Pro}_{\text{vis}}$ \cite{yue2024mmmu} while still improving accuracy by \textbf{6.7\%}.
Similarly, applying ChainV to Qwen3-VL-Thinking 8B \cite{yang2025qwen3} yields an average \textbf{34.1\%} latency reduction and a \textbf{3.4\%} accuracy gain across six benchmarks, demonstrating both its effectiveness and general compatibility with different multimodal reasoning models.

Our main contributions are summarized as follows:
\begin{itemize}
\item We introduce a training-free multimodal reasoning framework dubbed ChainV. It explores answer-aware visual hint guidance for efficient Chain-of-Thought inference.
\item  Particularly, we introduce a coarse-to-fine strategy to extract visual assistant regions aligned with the reasoning, along with a reliability judgment. Bernoulli-based scheduler to probabilistically inject visual hints, effectively reducing reflection loop and shortening the reasoning trace.
\item Our ChainV enables various multimodal reasoning models to \textit{think faster} and \textit{answer better}, achieving simultaneous gains in efficiency and accuracy on mathematical, logical, and scientific vision-language benchmarks.
\end{itemize} 

\begin{figure*}[t]
    \centering
    \includegraphics[width=1 \textwidth]{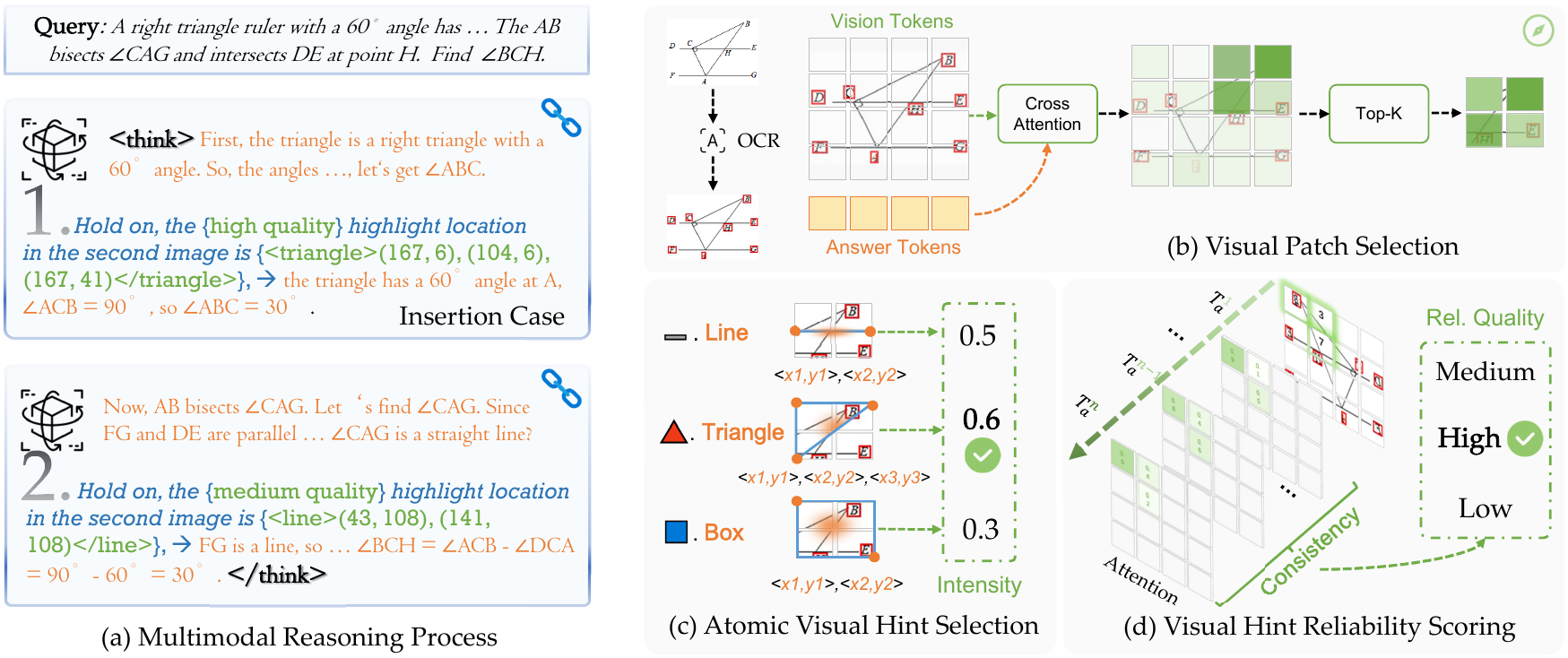}
    \vspace{-4mm}
    \caption{\textbf{The pipeline of ChainV.} In Figure (a), a multimodal reasoning model is solving a mathematical problem, during which our ChainV is invoked twice. Figure (b-d) shows the detailed process of ChainV, where the output is a visual hint annotated with coordinates.
}
    \vspace{-2mm}
    \label{fig:main}
\end{figure*}

\section{Related Work}

\subsection{Multimodal Chain-of-Thought Reasoning}
Multimodal reasoning \cite{meng2023chain, hu2024visual, zhang2025autov, liu2025visual, thawakar2025llamav, zhang2024improve, rafailov2023direct, xiaomi2025mimo, team2025kwai, yang2025qwen3} is a core component of AI models, enabling human-like, step-by-step inference across multimodal contexts (\eg, images and text). Over time, it has evolved through three paradigms: (1) Explicit modular reasoning, where models like Chain-of-Image \cite{meng2023chain} and Visual SketchPad \cite{hu2024visual} externalize reasoning via visual synthesis or sketching; (2) Supervised chain-of-thought reasoning, as in LLaVA-CoT \cite{xu2024llava} and LlamaV-o1 \cite{thawakar2025llamav}, which employ structured templates and step-wise supervision; and (3) Reinforcement learning optimization, adopted by LLaVA-Reasoner \cite{zhang2024improve}, MiMo-VL-RL \cite{xiaomi2025mimo}, Keye-VL-Thinking \cite{team2025kwai}, and Qwen3-VL-Thinking \cite{yang2025qwen3}.
Our work builds upon this final line, targeting the redundancy in implicit adaptive-length.

\subsection{Visual Injection for Chain-of-Thought}
Visual injection is a technique for enhancing CoT reasoning in multimodal contexts. Prior works, such as visual CoT models \cite{shao2024visual, meng2023chain}, rely on pre-selected visual cues to ground reasoning steps in relevant evidence. More recent approaches, including Look-Back \cite{yang2025look}, guide MLLMs to autonomously re-focus on visual inputs during reasoning, leveraging their intrinsic visual fusion capabilities by finetuning. MINT-CoT \cite{chen2025mint} advances in this direction by interleaving visual tokens into mathematical Chain-of-Thought, dynamically selecting relevant regions at the token level, and requiring a dedicated dataset for training. Unlike these training-intensive methods, ChainV targets training-free visual reasoning and optimizes efficiency during inference. We introduce a new category of visual injection, termed visual hints, which insert diverse atomic visual hints in an answer-aware manner with varying confidence levels.

\subsection{Efficiency in Chain-of-Thought Reasoning}
Standard CoT methods are computationally intensive due to the lengthy intermediate reasoning steps they generate \cite{guo2025deepseek, xiaomi2025mimo, yang2025qwen3}. Recent efforts, such as Chain-of-Draft \cite{xu2025chain}, mitigate this issue, where each reasoning step is constrained to only a few words to reduce token usage. Similarly, AlphaOne \cite{zhang2025alphaone} dynamically transitions from slow to fast reasoning through an $\alpha$-moment mechanism and stochastic transition tokens, deterministically terminating slow thinking for efficient output.
However, these efficiency-oriented paradigms remain limited to the textual domain, and no corresponding approaches have been explored for multimodal reasoning to date. To this end, we present ChainV, which optimizes multimodal reasoning efficiency in a training-free manner by injecting adaptive visual hints.

\section{Proposed Approach: \chainvexp}

\label{sec:method}
\subsection{Preliminary: Multimodal Reasoning Model}
Multimodal reasoning models process visual and textual inputs and produce a step-by-step chain-of-thought reasoning output, divided into two parts by the \texttt{</think>} token. 

Formally, for an original input image $\vx_v$, the vision embedding tokens $\mH_v \in \mathbb{R}^{L_v\times D}$ can be computed as
\begin{equation} \label{eq:1}
    \mH_v := \mW \mZ_v,
\end{equation}
where $L_v$ denotes the number of tokens, and $D$ is the dimension. $\mZ_v$ is the visual feature provided by visual encoder $\mZ_v := g(\vx_v)$, and $\mW$ is the projection that maps $\mZ_v$ to vision embedding tokens $\mH_v$. In this work, we additionally incorporate an OCR-recognized image \cite{cui2025paddleocr}, denoted as $\vx_o$, referred to as the visual assistant. The same embedding process is applied to obtain its corresponding tokens $\mH_o$.

Next, the visual tokens mentioned above, along with the query embedding $\mH_q$ obtained via the tokenizer, are fed into the language model to produce the thinking steps $\sT$,
\begin{equation}
    \sT := f_{\text{LM}}(\mH_q, [\mH_v; \mH_o]).
\end{equation}
$[\cdot;\cdot]$ denotes token concatenation, and $f_{\text{LM}}$ represents the language model that autoregressively generates the chain-of-thought reasoning output $\sT = \{\mT^{(1)}, \mT^{(2)}, \dots, \mT^{(n)}\}$ until the end-of-thought token is produced, and each $\mT^{(i)} \in \mathbb{R}^D$ is the embedding of the $i$-th generated answer token.

\subsection{Visual Patch Selection}
Previous chain-of-thought approaches in multimodal reasoning models primarily rely on textual reasoning, lacking explicit grounding in visual cues. This limitation hampers their ability to handle tasks requiring precise visual understanding, such as mathematical and geometric reasoning.

To address this gap, we introduce a mechanism that dynamically identifies visual patches most relevant to the current reasoning process, enabling the model to establish a tight \textit{see-and-think} correspondence throughout reasoning. Specifically, for each generated thinking token $\mT^{(i)}$, we compute its attention intensity with the auxiliary visual assistant embedding $\mH_o\in \mathbb{R}^{L_v\times D}$, measured by
\begin{equation}
    \mA^{(i)}_o := \mT^{(i)} (\mH_o)^{\top}.
\end{equation}
Then, to obtain a stable and reliable estimation of visual relevance across all recent thinking tokens, we compute the mean of their token-wise attention responses:
\begin{equation}
    \mA := \frac{1}{N_a} \sum_{i=1}^{N_a} \mA^{(i)}_o,
\end{equation}
where $N_a \ll n$ denotes the total number of recently generated thinking tokens within the current reasoning sentence, which \textit{preserves timeliness} while saving computation.

We then select the top-$k$ visual tokens according to the averaged attention distribution:
\begin{equation}
    \sI := \mbox{TopK}(\mA, k), \quad k\in\{3, \dots, L_v\}
\end{equation}
where $\sI$ denotes the indices of selected visual tokens, and $k=3$ in our implementation corresponds to the most salient visual regions that contribute to the model's reasoning process at the current step. With attention signals from the answer, \chainvexp initially identifies coarse-grained regions that are visually relevant to the current reasoning step.

\subsection{Atomic Visual Hint Selection}
While the above selection provides guidance, it is inherently restricted to patch regions, which fail to capture the fine-grained structures necessary for precise reasoning. 
Therefore, we decompose the coarse regions into fine-grained \textit{atomic visual hints}, with three types of elementary geometric units: \textbf{lines}, \textbf{triangles}, and \textbf{boxes} that provide more interpretable and semantically aligned hints.

\textbf{Line.} 
We first convert the selected patch indices $\sI$ into spatial center coordinates $\sP$. Then construct a line hint by applying principal component analysis (PCA)~\cite{wold1987principal} to extract the dominant direction:
\begin{equation}
    \mathcal{D} := \operatorname*{eigvec}_{\max}\!\big(\mbox{Cov}(\sP)\big),
\end{equation}
where $\mbox{Cov}(\cdot)$ is the covariance matrix and $\operatorname*{eigvec}_{\max}(\cdot)$ returns the eigenvector of the largest eigenvalue. Spatial centers are projected onto $\mathcal{D}$, and the two extreme points along this direction form the atomic line hint segment.

\textbf{Triangle.}
Based on the boundary coordinates of the selected patches, we construct the triangular atomic hint by locating the largest adjacent triangle within their geometric boundary. This atomic unit is represented by three vertices.

\textbf{Box.}
With the spatial layout of the selected patches, we compute the smallest bounding rectangle that encloses their spatial extent.  
This atomic box hint provides a coarse yet complete localization of the attended visual region.

Once we construct the atomic visual hints, we select one based on the average attention intensity within the corresponding region during the current reasoning step. Formally, given an atomic visual hint $\mathcal{R}$, we map the region to the pixel grid of the original image. A mask $\mM_\mathcal{R}$ is generated to indicate which pixels fall inside $\mathcal{R}$. The average attention within the region is computed as
\begin{equation}
\bar{\a}_\mathcal{R} := \frac{1}{|\mM_\mathcal{R}|} \sum_{(x,y) \in \mM_\mathcal{R}} \mA_{\text{pixel}}(x, y),
\end{equation}
where $\mA_{\text{pixel}}$ is the pixel-level attention map obtained by appropriately mapping token-level attention to the image, and $|\mM_\mathcal{R}|$ is the number of pixels within the region.

\subsection{Visual Hint Reliability Scoring}
Selecting the atomic visual hint with the highest response intensity raises questions about its reliability in guiding reasoning. A high similarity sum does not ensure stable attention across individual answer tokens. To address this, we propose a consistency-aware scoring mechanism designed to assign a reliable score to the hint, enabling the reasoning process to depend more effectively on the visual hint.

In particular, we employ the consistency of visual attention across different answer tokens within the visual hint region as an indicator of that region reliability.
Firstly, for the $i$th answer token, we normalize its attention over the selected region $\mathcal{R}$ as
\begin{equation}
    \hat{\a}^{(i)}_{\mathcal{R}} := 
    \frac{\mA_{\text{pixel}}^{(i)}(x, y)}
         {\sum_{(x',y') \in \mM_\mathcal{R}} \mA_{\text{pixel}}^{(i)}(x',y') + \varepsilon},
\end{equation}
where $\mA_{\text{pixel}}^{(i)}(x, y)$ is the pixel-level attention of the $i$th answer token, and $\varepsilon$ is a small constant.

We then adopt the widely-used Pearson correlation \cite{benesty2009pearson} $\rho_\mathrm{p}$ to measure the pairwise consistency between attention distributions, and the consistency score is defined as
\begin{equation}
    \mathcal{C}_{\mathcal{R}} := \sum_{i<j}\rho_\mathrm{p}(\hat{\a}^{(i)}, \, \hat{\a}^{(j)}),
\end{equation}
where
\begin{align} \label{eq:pearson}
\rho_\mathrm{p} (\u, \v)
&:= \frac{\mbox{Cov}(\u, \v)}{\mbox{Std}(\u)\,\mbox{Std}(\v)} \notag \\
&= \frac{\sum_{i=1}^C (u_i - \bar{u})(v_i - \bar{v})}
{\sqrt{\sum_{i=1}^C (u_i - \bar{u})^2
       \sum_{i=1}^C (v_i - \bar{v})^2}},
\end{align}
where $\mbox{Cov}(\u, \v)$ is the covariance of $\u$ and $\v$, $\bar{u}$ and $\mbox{Std}(\u)$ denote the mean and standard deviation of $\u$, respectively.

\subsection{Visual Hint Insertion}
After computing the precise consistency scores for the three atomic visual hints, we rank them and assign them labels of \textbf{high}, \textbf{medium}, or \textbf{low}. Eventually, we select the hint with the highest average attention intensity as the final visual hint. The resulting template $\mathcal{V}$ contains two \textit{variables}: the spatial coordinates of the hint and the score of reliability, as detailed below:
\[
\mathcal{V} := \textit{The } \textcolor{kaiming-green}{\textit{\{quality label\}}} \textit{ highlight location in the}
\]
\[
\textit{second image is } \textcolor{kaiming-green}{\textit{\{\textless hint type\textgreater hint location \textless /hint type\textgreater\}}} \textit{.}
\]
With the content $\mathcal{V}$ ready for insertion, our ChainV employs a Bernoulli stochastic process \cite{medhi1994stochastic}, using a trigger word at the start to interrupt and guide the reasoning. In our experiments, the trigger word is consistently set to “\textit{Hold on}.” Inspired by \cite{zhang2025alphaone, huang2025efficient}, the $\bm{p}_\text{Hold on}$ is decided by a linear annealing scheduling function, concerning the length of generated output tokens, indicating a fast-to-slow thinking strategy.

\section{Experiments}
\definecolor{mygray}{gray}{.95}
\begin{table*}[ht!]
\caption{\textbf{Comparison of \chainvexp with other training-free reasoning paradigms across various advanced models.} The benchmarks include challenging tasks in mathematics, reasoning, and science. \textbf{Time} refers to the average inference time per sample (in seconds). Method efficiency is measured by acc/s (accuracy per second), which is presented in the last column.}
\label{table:main_results}
\centering
\renewcommand{\arraystretch}{1.2} 
\setlength{\tabcolsep}{1.2mm}
\resizebox{\textwidth}{!}{
    \begin{tabular}{l|llllllllllll|ll}
    \toprule[0.95pt]
    \multirow{3}{*}[-1.0ex]{\textbf{Method}} 
    &
    \multicolumn{6}{c}{\scshape\textbf{Mathematical}}
    & \multicolumn{4}{c}{\scshape\textbf{Reasoning}}
    & \multicolumn{2}{c|}{\scshape\textbf{Science}}
    & \multicolumn{2}{c}{\multirow{2}{*}[-1.0ex]{\textbf{Avg.}}} \\ 
    \cmidrule(lr){2-7}
    \cmidrule(lr){8-11}
    \cmidrule(lr){12-13}
    &
    \multicolumn{2}{c}{\textbf{$\text{MathVista}_{\text{mini}}$}}
    & \multicolumn{2}{c}{\textbf{MathVision}}
    & \multicolumn{2}{c}{\textbf{WeMath}}
    & \multicolumn{2}{c}{\textbf{$\text{MMMU Pro}_{\text{vis}}$}}
    & \multicolumn{2}{c}{\textbf{LogicVista}}
    & \multicolumn{2}{c|}{\textbf{Olympiad}}
    & & \\   
    \cmidrule(lr){2-3}
    \cmidrule(lr){4-5}
    \cmidrule(lr){6-7}
    \cmidrule(lr){8-9}
    \cmidrule(lr){10-11}
    \cmidrule(lr){12-13}
    \cmidrule(lr){14-15}   
    & 
    Acc. $\uparrow$ & Time $\downarrow$ &
    Acc. $\uparrow$ & Time $\downarrow$ &
    Acc. $\uparrow$ & Time $\downarrow$ &
    Acc. $\uparrow$ & Time $\downarrow$ &
    Acc. $\uparrow$ & Time $\downarrow$ &
    Acc. $\uparrow$ & Time $\downarrow$ &
    Acc. $\uparrow$ & Time $\downarrow$ \\
    \midrule[0.6pt]
    \multicolumn{14}{c}{\textit{\textbf{MiMo-VL-RL}} (7B)}\\
    \baselineexp & 81.4 & \textcolor{kaiming-green}{3.5} & {59.9} & \textcolor{kaiming-green}{5.1} & 66.0 & \textcolor{kaiming-green}{1.6} & 40.2 & \textcolor{kaiming-green}{4.2} & 61.4 & \textcolor{kaiming-green}{6.1} & 57.4 & \textcolor{kaiming-green}{5.8} & 61.1 & \textcolor{kaiming-green}{4.4} \\ 
    
    \Noexp & 76.0 & \textcolor{kaiming-green}{3.0} & 52.1 & \textcolor{kaiming-green}{4.9} & 58.3 & \textcolor{kaiming-green}{1.9} & 37.6 & \textcolor{kaiming-green}{4.0} & 53.3 & \textcolor{kaiming-green}{4.7} & 50.4 & \textcolor{kaiming-green}{5.0} & 54.6 & \textcolor{kaiming-green}{3.9} \\ 
    
    \codexp & 82.0 & \textcolor{kaiming-green}{2.5} & 57.9 & \textcolor{kaiming-green}{5.0} & 63.9 & \textcolor{kaiming-green}{2.1} & 40.4 & \textcolor{kaiming-green}{\textbf{1.0}} & 60.3 & \textcolor{kaiming-green}{\textbf{3.7}} & {57.4} & \textcolor{kaiming-green}{5.7} & 60.3 & \textcolor{kaiming-green}{3.3} \\ 

    \alphaexp & {82.0} & \textcolor{kaiming-green}{\textbf{1.5}} & 58.1 & \textcolor{kaiming-green}{{4.2}} & {66.0} & \textcolor{kaiming-green}{{1.5}} & {41.3} & \textcolor{kaiming-green}{3.7} & {63.3} & \textcolor{kaiming-green}{{3.8}} & 56.2 & \textcolor{kaiming-green}{{4.5}} & 61.1 & \textcolor{kaiming-green}{3.2} \\ 

    \rowcolor{mygray}
    \chainvexp & \textbf{83.3}$_{\text{\textcolor{runpei-orange}{\textbf{+1.9}}}}$ & \textcolor{kaiming-green}{{1.7}} &\textbf{61.1}$_{\text{\textcolor{runpei-orange}{\textbf{+1.2}}}}$ & \textcolor{kaiming-green}{\textbf{4.0}}& \textbf{68.4}$_{\text{\textcolor{runpei-orange}{\textbf{+2.4}}}}$ & \textcolor{kaiming-green}{\textbf{1.5}} & \textbf{43.0}$_{\text{\textcolor{runpei-orange}{\textbf{+2.8}}}}$ & \textcolor{kaiming-green}{{3.3}} & \textbf{63.9}$_{\text{\textcolor{runpei-orange}{\textbf{+2.5}}}}$ & \textcolor{kaiming-green}{4.1} & \textbf{59.4}$_{\text{\textcolor{runpei-orange}{\textbf{+2.0}}}}$ & \textcolor{kaiming-green}{\textbf{4.1}} & \textbf{63.2}$_{\text{\textcolor{runpei-orange}{\textbf{+2.1}}}}$ & \textcolor{kaiming-green}{\textbf{3.1}}$_{\text{\textcolor{kaiming-green}{$\downarrow$ 30\%}}}$  \\ 
    
    \midrule[0.6pt]
    \multicolumn{14}{c}{\textit{\textbf{Keye-VL-Thinking}} (8B)}\\
    \baselineexp & 75.2 & \textcolor{kaiming-green}{3.9} & 42.5 & \textcolor{kaiming-green}{5.7} & 58.8 & \textcolor{kaiming-green}{1.4} & 32.9 & \textcolor{kaiming-green}{4.7} & 50.2 & \textcolor{kaiming-green}{5.9} & 45.2 & \textcolor{kaiming-green}{5.5} & 50.8 & \textcolor{kaiming-green}{4.5}  \\ 
    
    \Noexp & 70.8 & \textcolor{kaiming-green}{3.1} & 41.0 & \textcolor{kaiming-green}{5.0} & 55.8 & \textcolor{kaiming-green}{1.6} & 31.6 & \textcolor{kaiming-green}{4.4} & 48.4 & \textcolor{kaiming-green}{5.7} & 43.0 & \textcolor{kaiming-green}{4.5} & 48.4 & \textcolor{kaiming-green}{4.1} \\ 
    
    \codexp & 76.9 & \textcolor{kaiming-green}{3.3} & 40.7 & \textcolor{kaiming-green}{5.2} & 59.4 & \textcolor{kaiming-green}{1.8} & 34.5 & \textcolor{kaiming-green}{{2.5}} & 50.6 & \textcolor{kaiming-green}{{4.3}} & 46.0 & \textcolor{kaiming-green}{5.0} & 51.4 & \textcolor{kaiming-green}{3.7} \\ 

    \alphaexp & {77.3} & \textcolor{kaiming-green}{{3.0}} & {42.9} & \textcolor{kaiming-green}{{4.8}} & {60.1} & \textcolor{kaiming-green}{{1.3}} & {34.9} & \textcolor{kaiming-green}{\textbf{2.3}} & {54.0} & \textcolor{kaiming-green}{5.0} & {46.2} & \textcolor{kaiming-green}{{4.3}} & 52.6 & \textcolor{kaiming-green}{3.5} \\ 

    \rowcolor{mygray}
    \chainvexp & \textbf{77.9}$_{\text{\textcolor{runpei-orange}{\textbf{+2.7}}}}$ & \textcolor{kaiming-green}{\textbf{2.5}} & \textbf{43.9}$_{\text{\textcolor{runpei-orange}{\textbf{+1.4}}}}$ & \textcolor{kaiming-green}{\textbf{4.2}}  & \textbf{61.9}$_{\text{\textcolor{runpei-orange}{\textbf{+3.1}}}}$ & \textcolor{kaiming-green}{\textbf{1.1}} & \textbf{35.1}$_{\text{\textcolor{runpei-orange}{\textbf{+2.2}}}}$ & \textcolor{kaiming-green}{2.6} & \textbf{54.6}$_{\text{\textcolor{runpei-orange}{\textbf{+4.4}}}}$ & \textcolor{kaiming-green}{\textbf{3.6}} & \textbf{47.1}$_{\text{\textcolor{runpei-orange}{\textbf{+1.9}}}}$ & \textcolor{kaiming-green}{\textbf{4.3}} & \textbf{53.4}$_{\text{\textcolor{runpei-orange}{\textbf{+2.6}}}}$ & \textcolor{kaiming-green}{\textbf{3.1}}$_{\text{\textcolor{kaiming-green}{$\downarrow$ 31\%}}}$  \\ 

    \midrule[0.6pt]
    \multicolumn{14}{c}{\textit{\textbf{Qwen3-VL-Thinking}} (2B)}\\
    \baselineexp & 73.1 & \textcolor{kaiming-green}{2.8} & 44.7 & \textcolor{kaiming-green}{3.7} & 50.4 & \textcolor{kaiming-green}{2.9} & 27.9 & \textcolor{kaiming-green}{3.1} & 30.4 & \textcolor{kaiming-green}{5.9} & 44.8 & \textcolor{kaiming-green}{3.8} & 45.2 & \textcolor{kaiming-green}{3.7} \\ 
    
    \Noexp & 71.8 & \textcolor{kaiming-green}{1.4} & 42.8 & \textcolor{kaiming-green}{2.9} & 40.2 & \textcolor{kaiming-green}{1.8} & 23.3 & \textcolor{kaiming-green}{2.8} & 28.2 & \textcolor{kaiming-green}{4.0} & 40.8 & \textcolor{kaiming-green}{\textbf{2.3}} & 41.2 & \textcolor{kaiming-green}{\textbf{2.5}}\\
    
    \codexp & 74.9 & \textcolor{kaiming-green}{1.6} & 43.0 & \textcolor{kaiming-green}{2.9} & 50.4 & \textcolor{kaiming-green}{2.0} & 30.4 & \textcolor{kaiming-green}{\textbf{2.4}} & 32.3 & \textcolor{kaiming-green}{4.2} & 44.2 & \textcolor{kaiming-green}{3.1} & 45.9 & \textcolor{kaiming-green}{2.7}\\

    \alphaexp & 76.0 & \textcolor{kaiming-green}{\textbf{1.3}} & 44.9 & \textcolor{kaiming-green}{3.1} & 51.1 & \textcolor{kaiming-green}{1.8} & 31.9 & \textcolor{kaiming-green}{2.6} & 35.5 & \textcolor{kaiming-green}{\textbf{3.7}} & 45.8 & \textcolor{kaiming-green}{3.2} & 47.5 & \textcolor{kaiming-green}{2.6}\\

    \rowcolor{mygray}
    \chainvexp & \textbf{76.9}$_{\text{\textcolor{runpei-orange}{\textbf{+3.8}}}}$ & \textcolor{kaiming-green}{1.6} & \textbf{45.7}$_{\text{\textcolor{runpei-orange}{\textbf{+1.0}}}}$ & \textcolor{kaiming-green}{\textbf{2.7}} & 52.3$_{\text{\textcolor{runpei-orange}{\textbf{+1.9}}}}$ & \textcolor{kaiming-green}{\textbf{1.6}} & \textbf{33.2}$_{\text{\textcolor{runpei-orange}{\textbf{+5.3}}}}$ & \textcolor{kaiming-green}{2.6} & \textbf{37.7}$_{\text{\textcolor{runpei-orange}{\textbf{+7.3}}}}$ & \textcolor{kaiming-green}{4.1} & \textbf{46.2}$_{\text{\textcolor{runpei-orange}{\textbf{+1.4}}}}$ & \textcolor{kaiming-green}{2.8} & \textbf{48.7}$_{\text{\textcolor{runpei-orange}{\textbf{+3.5}}}}$ & \textcolor{kaiming-green}{2.6}$_{\text{\textcolor{kaiming-green}{$\downarrow$ 30\%}}}$ \\ 

    \midrule[0.6pt]
    \multicolumn{14}{c}{\textit{\textbf{Qwen3-VL-Thinking}} (8B)}\\
    \baselineexp & 81.0 & \textcolor{kaiming-green}{3.7} & 61.4 & \textcolor{kaiming-green}{3.7} & 58.1 & \textcolor{kaiming-green}{3.0} & 39.7 & \textcolor{kaiming-green}{3.6} & 55.8 & \textcolor{kaiming-green}{6.1} & 54.1 & \textcolor{kaiming-green}{4.7} & 58.5 & \textcolor{kaiming-green}{4.1} \\ 
    
    \Noexp & 78.2 & \textcolor{kaiming-green}{2.0} & 57.9 & \textcolor{kaiming-green}{2.8} & 55.8 & \textcolor{kaiming-green}{2.6} & 37.0 & \textcolor{kaiming-green}{2.6} & 53.6 & \textcolor{kaiming-green}{5.0} & 51.4 & \textcolor{kaiming-green}{3.0} & 55.7 & \textcolor{kaiming-green}{3.0}\\
    
    \codexp & 81.9 & \textcolor{kaiming-green}{2.0} & 61.0 & \textcolor{kaiming-green}{3.0} & 59.3 & \textcolor{kaiming-green}{2.3} & 40.0 & \textcolor{kaiming-green}{\textbf{2.1}} & 56.6 & \textcolor{kaiming-green}{\textbf{4.7}} & 54.0 & \textcolor{kaiming-green}{3.3} & 58.8 & \textcolor{kaiming-green}{2.9}\\

    \alphaexp & 81.9 & \textcolor{kaiming-green}{1.8} & 62.4 & \textcolor{kaiming-green}{2.5} & 59.7 & \textcolor{kaiming-green}{2.4} & 40.2 & \textcolor{kaiming-green}{2.7} & 58.3 & \textcolor{kaiming-green}{5.3} & 54.6 & \textcolor{kaiming-green}{2.9} & 59.5 & \textcolor{kaiming-green}{2.9}\\

    \rowcolor{mygray}
    \chainvexp & \textbf{83.0}$_{\text{\textcolor{runpei-orange}{\textbf{+2.0}}}}$ & \textcolor{kaiming-green}{\textbf{1.5}} & \textbf{62.9}$_{\text{\textcolor{runpei-orange}{\textbf{+1.5}}}}$ & \textcolor{kaiming-green}{\textbf{2.2}} & \textbf{60.7}$_{\text{\textcolor{runpei-orange}{\textbf{+2.6}}}}$ & \textcolor{kaiming-green}{\textbf{2.1}} & \textbf{41.6}$_{\text{\textcolor{runpei-orange}{\textbf{+1.9}}}}$ & \textcolor{kaiming-green}{2.8} & \textbf{59.0}$_{\text{\textcolor{runpei-orange}{\textbf{+3.2}}}}$ & \textcolor{kaiming-green}{5.0} & \textbf{55.9}$_{\text{\textcolor{runpei-orange}{\textbf{+1.8}}}}$ & \textcolor{kaiming-green}{\textbf{2.9}} & \textbf{60.5}$_{\text{\textcolor{runpei-orange}{\textbf{+2.0}}}}$ & \textcolor{kaiming-green}{\textbf{2.7}}$_{\text{\textcolor{kaiming-green}{$\downarrow$ 34\%}}}$ \\ 
    
    \bottomrule[0.95pt]
    \end{tabular}
}
\end{table*}
\subsection{Experimental Settings}

\paragraph{Implementation Details.}
All inference experiments are conducted on $8 \times$ NVIDIA A100 GPUs. To ensure reproducibility, we fix all random seeds and maintain consistent GPU memory utilization and inference batch sizes across experiments on the same model, preventing fluctuations in inference latency. Details can be found in Appendix \ref{sec:details}.

\paragraph{Benchmarks.} 
We evaluate our method on six challenging multimodal reasoning tasks, including MathVista \cite{lu2023mathvista}, MathVision \cite{wang2024measuring}, WeMath \cite{qiao2024we}, MMMU Pro \cite{yue2024mmmu}, LogicVista \cite{xiao2024logicvista}, and OlympiadBench \cite{he2024olympiadbench}. In addition, we further evaluate it on general visual understanding tasks, including VStar \cite{wu2024v}, CVBench \cite{tong2024cambrian}, ConBench \cite{zhang2024unveiling}, ChartVQA \cite{masry2022chartqa}, SEED-Bench \cite{li2024seed}, and ScreenSpot \cite{jurmu2008screenspot}, resulting in a total of \textbf{twelve benchmarks}.
Their characteristics and categories are detailed in Appendix \ref{sec:ben_details}.

\paragraph{Models.} 
We apply ChainV to three open-source multimodal reasoning models: MiMo-VL-RL-7B \cite{xiaomi2025mimo}, Keye-VL-8B-Thinking \cite{team2025kwai}, and the latest Qwen3-VL-2B/8B-Thinking \cite{yang2025qwen3}, which vary significantly in model size.

\paragraph{Baselines.} 
We evaluate ChainV against four training-free baselines: \baselineexp, which automatically switches between slow and fast thinking; \Noexp, which disables slow thinking; \codexp paradigm \cite{xu2025chain}; and \alphaexp schedule \cite{zhang2025alphaone}.

\definecolor{mygray}{gray}{.95}
\begin{table*}[ht!]
\caption{\textbf{Comparison of \chainvexp with other reasoning paradigms on MiMo-VL-RL.} The benchmarks cover vision-centric understanding, optical character recognition, and GUI understanding. For better distinction, the time values are kept to \textit{two decimal places}.
}
\label{table:general_results}
\centering
\renewcommand{\arraystretch}{1.25} 
\setlength{\tabcolsep}{1.2mm}
\resizebox{\textwidth}{!}{
    \begin{tabular}{l|llllllllllll|ll}
    \toprule[0.95pt]
    \multirow{3}{*}[-1.0ex]{\textbf{Method}} 
    &
    \multicolumn{6}{c}{\scshape\textbf{Vision-Centric}}
    & \multicolumn{4}{c}{\scshape\textbf{OCR}}
    & \multicolumn{2}{c|}{\scshape\textbf{GUI}}
    & \multicolumn{2}{c}{\multirow{2}{*}[-1.0ex]{\textbf{Avg.}}} \\ 
    \cmidrule(lr){2-7}
    \cmidrule(lr){8-11}
    \cmidrule(lr){12-13}
    &
    \multicolumn{2}{c}{\textbf{VStar}}
    & \multicolumn{2}{c}{\textbf{CVBench}}
    & \multicolumn{2}{c}{\textbf{ConBench}}
    & \multicolumn{2}{c}{\textbf{ChartVQA}}
    & \multicolumn{2}{c}{\textbf{SEED-Bench}}
    & \multicolumn{2}{c|}{\textbf{ScreenSpot}}
    & & \\   
    \cmidrule(lr){2-3}
    \cmidrule(lr){4-5}
    \cmidrule(lr){6-7}
    \cmidrule(lr){8-9}
    \cmidrule(lr){10-11}
    \cmidrule(lr){12-13}
    \cmidrule(lr){14-15}   
    & 
    Acc. $\uparrow$ & Time $\downarrow$ &
    Acc. $\uparrow$ & Time $\downarrow$ &
    Acc. $\uparrow$ & Time $\downarrow$ &
    Acc. $\uparrow$ & Time $\downarrow$ &
    Acc. $\uparrow$ & Time $\downarrow$ &
    Acc. $\uparrow$ & Time $\downarrow$ &
    Acc. $\uparrow$ & Time $\downarrow$ \\
    \midrule[0.6pt]
    \baselineexp & 81.7 & \textcolor{kaiming-green}{1.34} & 82.1 & \textcolor{kaiming-green}{0.42} & 26.3 & \textcolor{kaiming-green}{1.35} & 91.7 & \textcolor{kaiming-green}{1.39} & 72.9 & \textcolor{kaiming-green}{0.52} & 87.1 & \textcolor{kaiming-green}{0.33} & 73.6 & \textcolor{kaiming-green}{0.89}\\ 
    
    \Noexp & 82.2 & \textcolor{kaiming-green}{\textbf{1.21}} & 82.0 & \textcolor{kaiming-green}{0.53} & 27.8 & \textcolor{kaiming-green}{1.01} & 92.9 & \textcolor{kaiming-green}{\textbf{1.22}} & 69.9 & \textcolor{kaiming-green}{0.51} & 87.2 & \textcolor{kaiming-green}{0.30} & 73.7 & \textcolor{kaiming-green}{0.80}\\ 
    
    \codexp & 80.6 & \textcolor{kaiming-green}{1.46} & 77.9 & \textcolor{kaiming-green}{0.73} & 13.2 & \textcolor{kaiming-green}{1.03} & 91.0 & \textcolor{kaiming-green}{1.67} & 65.2 & \textcolor{kaiming-green}{0.81} & 87.7 & \textcolor{kaiming-green}{0.29} & 69.3 & \textcolor{kaiming-green}{1.00}\\ 

    \alphaexp & 82.2 & \textcolor{kaiming-green}{1.28} & 82.4 & \textcolor{kaiming-green}{0.40} & 27.2 & \textcolor{kaiming-green}{1.12} & 92.1 & \textcolor{kaiming-green}{1.53} & 72.9 & \textcolor{kaiming-green}{\textbf{0.46}} & 87.6 & \textcolor{kaiming-green}{0.30} & 74.1 & \textcolor{kaiming-green}{0.85}\\ 

    \rowcolor{mygray}
    \chainvexp & \textbf{82.7}$_{\text{\textcolor{runpei-orange}{\textbf{+1.0}}}}$ & \textcolor{kaiming-green}{{1.26}} &\textbf{83.3}$_{\text{\textcolor{runpei-orange}{\textbf{+1.2}}}}$ & \textcolor{kaiming-green}{\textbf{0.38}}& \textbf{29.0}$_{\text{\textcolor{runpei-orange}{\textbf{+2.7}}}}$ & \textcolor{kaiming-green}{\textbf{1.01}} & \textbf{92.9}$_{\text{\textcolor{runpei-orange}{\textbf{+1.2}}}}$ & \textcolor{kaiming-green}{{1.29}} & \textbf{74.9}$_{\text{\textcolor{runpei-orange}{\textbf{+2.0}}}}$ & \textcolor{kaiming-green}{0.55} & \textbf{88.2}$_{\text{\textcolor{runpei-orange}{\textbf{+1.1}}}}$ & \textcolor{kaiming-green}{\textbf{0.29}} & \textbf{75.2}$_{\text{\textcolor{runpei-orange}{\textbf{+1.6}}}}$ & \textcolor{kaiming-green}{\textbf{0.79}}$_{\text{\textcolor{kaiming-green}{$\downarrow$ 11\%}}}$  \\ 

    \bottomrule[0.95pt]
    \end{tabular}
}

\end{table*}
\subsection{Main Results}
\label{sec:exp}
As shown in Table~\ref{table:main_results}, ChainV consistently improves both the accuracy and inference efficiency across four representative multimodal reasoning models ($2$B$\sim$$8$B scale) and six challenging benchmarks spanning mathematics, logical reasoning, and scientific problem solving. Unlike prior efficiency-oriented decoding strategies that typically trade accuracy for latency reduction, ChainV achieves a favorable Pareto shift: it reduces inference time in all settings while maintaining or even boosting task performance.

\textbf{Accuracy Gains.}
On average, ChainV brings $+2.1 \sim 3.5$ absolute improvement in accuracy over the baseline models while outperforming other comparative methods. The gains are especially pronounced on benchmarks that urgently require multi-step symbolic reasoning, \emph{e.g.}, Qwen3-VL-Thinking 2B achieves improvements of $+3.8$ on MathVista and $+7.3$ on LogicVista when equipped with ChainV. Furthermore, ChainV is the only method that yields an average accuracy gain on the MiMo-VL-RL, whereas all other LLM-oriented approaches either degrade performance or show no improvement. For instance, although the state-of-the-art AlphaOne \cite{zhang2025alphaone} reduces latency, it fails to deliver accuracy gains on average. This confirms that the visual hint offers timely grounding to the intermediate reasoning trajectory, enabling more precise step-by-step reasoning.

\textbf{Latency Reduction.}
Here, the measured time refers to the Inference Time Latency (ITL), which includes tokenizer, prefill, decoding loop, and post-processing.
ChainV yields consistent reductions in inference time, with an average $30\% \sim 34\%$ relative speedup across all baseline models. Specifically, when equipped with ChainV, Qwen3-VL-Thinking 8B on MathVista reduces its per-sample inference time from $3.7$s to $1.5$s, corresponding to a $59.5\%$ reduction ($2.47\times$ speedup). The improvement mainly comes from (1) fewer “\textit{wait cycles}” during staged decoding and (2) a shorter final token trace enabled by early visual anchoring. Notably, even for the $2$B Qwen model with already low decoding overhead, the inference time still drops from $3.7$s to $2.6$s, corresponding to a $\sim30\%$ reduction.

\textbf{Robustness Beyond Models and Tasks.}
The efficacy of ChainV is not confined to a specific multimodal model architecture or training paradigm. Consistent improvements are observed on RL-enhanced models (MiMo-VL-RL), adaptive thinking models (Keye-VL-Thinking), and MoE-based Qwen3-VL variants, demonstrating robust cross-architecture performance. Beyond architectural diversity, ChainV exhibits strong cross-task robustness, yielding stable gains across math, logical reasoning, and science QA benchmarks. Additional experiments on general multimodal benchmarks\footnote{Here, the visual assistant is generated by a tiny object detector.}, reported in the Table \ref{table:general_results}, further validate this finding.
These results indicate that ChainV functions as a model-agnostic reasoning framework rather than an architecture- or task-specific optimization. The stronger gains over state-of-the-art methods on math-heavy datasets further support our hypothesis that visual priors are especially valuable when multi-step symbolic reasoning steps are explicitly grounded in visual evidence.

\section{Analysis}

\definecolor{mygraytext}{gray}{.75}
\newcommand{\cmark}{\ding{51}}%
\newcommand{\xmark}{\ding{55}}
\newcommand{\graytext}[1]{\textcolor{gray}{\raisebox{0.15ex}{#1}}}

\begin{table}[t]
  \caption{\textbf{Ablation of components in ChainV.} va: visual assistant, vp(2): visual patch selection in section 3.2, vh(3): atomic hint selection in section 3.3, re(4): reliability scoring in section 3.4.}
  \vspace{-1mm}
  \renewcommand{\arraystretch}{1.25}
  \setlength{\tabcolsep}{0.7mm}
  \label{tab:ab}
  \centering
  \begin{tabular}{p{0.9cm}|p{0.9cm}p{0.9cm}p{0.9cm}|cc|cc}
    \toprule
    \multirow{2}{*}{\textit{va}} &
    \multirow{2}{*}{\textit{vp}(2)} &
    \multirow{2}{*}{\textit{vh}(3)} &
    \multirow{2}{*}{\textit{re}(4)} &
    \multicolumn{2}{c|}{\textbf{$\text{MathVista}_{\text{mini}}$}} &
    \multicolumn{2}{c}{\textbf{Olympiad}} \\
    \cmidrule(lr){5-6} \cmidrule(lr){7-8}
     & & & & Acc. $\uparrow$ & Time $\downarrow$ & Acc. $\uparrow$ & Time $\downarrow$ \\
    \midrule
    \graytext{\xmark} & \graytext{\xmark} & \graytext{\xmark} & \graytext{\xmark} &
    81.4 & \textcolor{kaiming-green}{3.5} & 57.4 & \textcolor{kaiming-green}{5.8} \\
    
    \xmark & \cmark & \cmark & \cmark & 82.9 & \textcolor{kaiming-green}{1.6} & 58.8 & \textcolor{kaiming-green}{3.9}\\
    \midrule
    
    \cmark & \graytext{\xmark} & \graytext{\xmark} & \graytext{\xmark} & 81.9 & \textcolor{kaiming-green}{4.3} & 58.1 & \textcolor{kaiming-green}{6.6} \\
    
    \graytext{\cmark} & \cmark & \graytext{\xmark} & \graytext{\xmark} & 82.5 & \textcolor{kaiming-green}{1.5} & 58.7 & \textcolor{kaiming-green}{3.7} \\
    
    \graytext{\cmark} & \graytext{\cmark} & \cmark & \graytext{\xmark} & 83.0 & \textcolor{kaiming-green}{1.6} & 59.2 & \textcolor{kaiming-green}{4.0}\\
    
    \rowcolor{mygray}
    \graytext{\cmark} & \graytext{\cmark} & \graytext{\cmark} & \cmark & \textbf{83.3} & \textcolor{kaiming-green}{1.7} & \textbf{59.4} & \textcolor{kaiming-green}{4.1} \\
    \bottomrule
  \end{tabular}
  \vspace{-2mm}
\end{table}

\begin{figure*}[t]
    \centering
    \includegraphics[width=1.01\textwidth]{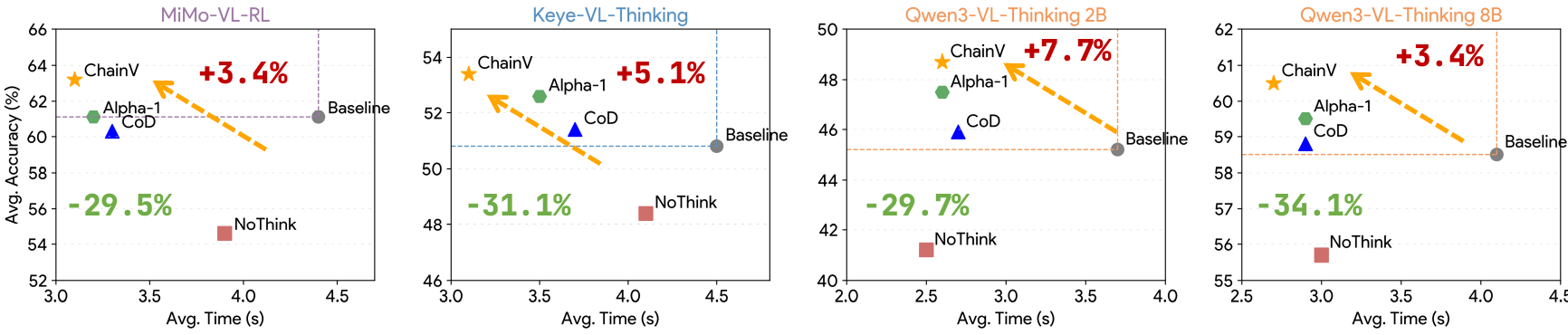}
    \vspace{-5mm}
    \caption{\textbf{Accuracy--Latency trade-off on multimodal reasoning models.} Green and red numbers indicate the reduction in inference time and the accuracy improvement achieved by ChainV compared to the baseline, respectively. Yellow arrows point toward better performance.}
    \vspace{-4mm}
    \label{fig:eff1}
\end{figure*}

\subsection{Effects of Components in ChainV}
To validate the effectiveness of each proposed component in ChainV, we perform ablation experiments, as shown in Table \ref{tab:ab}. The metrics include accuracy and wall-clock inference time. \textbf{+Visual Assistant.} The visual assistant essentially provides higher-quality visual hints generated by a lightweight OCR model. It directly improves reasoning accuracy; for example, it raises OlympiadBench accuracy by $+0.7$, but it also increases time cost by $0.8$s. Interestingly, comparing the 2nd and last rows shows that our ChainV maintains its benefits while mitigating costs: with only a $0.2$s increase, accuracy still improves by $+0.6$. \textbf{+Visual Patch Selection.} Performing visual patch localization based on the current reasoning step is fundamental and crucial. This module starts a visual interruption mechanism during text thinking, leading to $+0.6$ improvement on both benchmarks and, more importantly, a substantial reduction in time latency. The reason is that it reduces the number of redundant reflection loop. \textbf{+Atomic Visual Hint Selection.} After undergoing coarse-grained visual patch localization, we refine it using atomic visual hint, which leads to a noticeable boost in accuracy. However, constructing these hints adds some overhead, especially for line segments requiring PCA. The $0.3$s extra delay on the OlympiadBench likely comes from large-area hints enlarging the PCA matrices. \textbf{+Visual Hint Reliability Scoring.} After determining the final region, we provide the model with the reliability of this hint, allowing it to adaptively adjust its level of self-reflection. If the hint is highly reliable, little additional reflection is needed; otherwise, more reasoning is required. This yields a $+0.2\sim0.3$ improvement while adding only $0.1$s of delay on both benchmarks, thanks to the low computational cost of our Pearson correlation. 

Eventually, our full ChainV achieves an average significant improvement of $+2.0$ points and saves $1800$ms compared to the baseline on these two benchmarks.

\subsection{Efficiency Analysis}
To further examine the efficiency of our proposed method, we conduct a detailed analysis from two complementary perspectives: \textit{accuracy vs.~inference time} and \textit{accuracy vs.~output length}. These metrics reflect how effectively the ChainV balances reasoning quality and computational cost, as well as its advantages over existing efficient approaches.

\textbf{Accuracy vs. Latency.}
We have already reported the detailed accuracy and inference time in the main table \ref{table:main_results}. To better illustrate the “\textit{faster and better}” property of our approach, we further visualize the results in Figure \ref{fig:eff1}. Across four evaluated reasoning models, ChainV achieves a clear Pareto improvement over other comparative methods, consistently shifting the accuracy–latency frontier upward and leftward. For instance, compared with the default reasoning mode, ChainV reduces average inference time by over $30\%$ while boosting relative performance by $3\%\sim 8\%$. On the Qwen3-VL-Thinking 8B, our method further surpasses AlphaOne \cite{zhang2025alphaone}, achieving relative $1.7\%$ higher accuracy and $200$ms faster inference on average. This demonstrates that, compared with pure textual prompts, our visual hint enables the model to reach more accurate conclusions with fewer iterative steps and lower computational overhead.

\begin{figure}[t]
    \centering
    \includegraphics[width=0.48\textwidth]{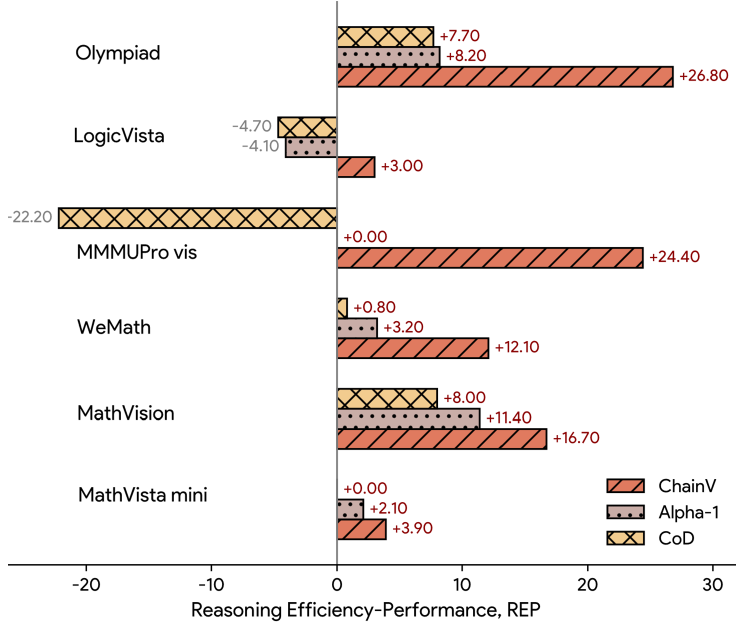}
    \vspace{-2mm}
    \caption{\textbf{Comparison of REP metric across six benchmarks}, based on the accuracy and the length of output tokens. Higher REP indicates a more favorable trade-off between reasoning accuracy and output efficiency. The evaluated model is MiMo-VL-RL 7B.
    }
    \vspace{-4mm}
    \label{fig:eff2}
\end{figure}

\textbf{Accuracy vs. Output Token Length.}
We then analyze the efficiency concerning answer accuracy and the number of generated output tokens. In current large-model deployment scenarios, the number of output tokens is directly tied to inference cost, making token compression a crucial factor for practical efficiency. Here, we adopt the REP metric proposed in \cite{zhang2025alphaone},  which is defined as follows:
\begin{equation}\label{eq:REP}
\mbox{REP}^{(i)}
:= (\mathcal{A}^{(i)} - \mathcal{A}_{\text{base}}) 
  \cdot \frac{\mathcal{T}_{\text{max}}}{\mathcal{T}^{(i)}},
\end{equation}
where $\mathcal{A}^{(i)}$ and $\mathcal{T}^{(i)}\gg n^{(i)}$ denote the accuracy and total number of output tokens of the $i$th method, respectively. $\mathcal{A}_{\text{base}}$ is the reasoning accuracy of the base model, and $\mathcal{T}_{\text{max}}$ is the maximum token length in setting, used for normalization. Higher REP indicates stronger performance with better reasoning efficiency. As illustrated in Figure~\ref{fig:eff2}, we compare the REP scores of CoD, AlphaOne, and ChainV across six reasoning benchmarks using the MiMo-VL-RL 7B model. ChainV consistently achieves the highest REP on all benchmarks, demonstrating a more favorable trade-off between reasoning performance and efficiency. 
In particular, ChainV surpasses CoD by $+19.10$ and AlphaOne by $+18.60$ on OlympiadBench, and further exceeds AlphaOne by $+24.40$ on $\text{MMMU Pro}_{\text{vis}}$. These results highlight that our ChainV is highly economical and serves as a more cost-effective and efficient multimodal reasoning plugin.

\subsection{Reliability Work Better as Numbers or Words?}
\begin{table}[t]
    \caption{\textbf{Ablation on reliability score representation.} A quantitative score (numbers) versus a qualitative description (words).}
  \vspace{-1mm}
  \renewcommand{\arraystretch}{1.2}
  \setlength{\tabcolsep}{0.95mm}
  \label{tab:reliability}
  \centering
  \begin{tabular}{p{1.5cm}p{1.5cm}|cc|cc}
    \toprule
    \multirow{2}{*}{\textit{numbers}} &
    \multirow{2}{*}{\textit{words}} &
    \multicolumn{2}{c|}{\textbf{$\text{MathVista}_{\text{mini}}$}} &
    \multicolumn{2}{c}{\textbf{Olympiad}} \\
    \cmidrule(lr){3-4} \cmidrule(lr){5-6}
     & & Acc. $\uparrow$ & Time $\downarrow$ & Acc. $\uparrow$ & Time $\downarrow$ \\
    \midrule
    \xmark & \xmark & 83.0 & \textcolor{kaiming-green}{1.6} & 59.2 & \textcolor{kaiming-green}{4.0} \\
    
    \cmark & \xmark & 82.8 & \textcolor{kaiming-green}{2.0} & 58.9 & \textcolor{kaiming-green}{4.2} \\

    \rowcolor{mygray}
    \xmark & \cmark & 83.3 & \textcolor{kaiming-green}{1.7} & 59.4 & \textcolor{kaiming-green}{4.1} \\
    
    \cmark & \cmark & 83.0 & \textcolor{kaiming-green}{1.9} & 58.8 & \textcolor{kaiming-green}{4.3}\\
    \bottomrule
  \end{tabular}
  \vspace{-3mm}
\end{table}

To investigate how the reliability score should be represented, we compare two strategies on MIMO-VL-RL: providing a quantitative score versus a qualitative description of hint reliability. In the quantitative setting, reliability is injected as a continuous scalar value, with its range indicating confidence. In contrast, the qualitative setting conveys reliability through natural language cues (\eg, high, medium).
As shown in Table~\ref{tab:reliability}, using qualitative descriptions consistently yields better reasoning accuracy while maintaining comparable or even lower latency. For instance, on MathVista, the qualitative form improves accuracy from $83.0$ to $83.3$ with reduced inference time. Similar trends are observed on the OlympiadBench. 
We hypothesize that, unlike spatial coordinates, where the model has already developed a quantitative understanding during pre-training and fine-tuning, the notion of confidence remains abstract and difficult to interpret numerically. Therefore, conveying reliability through relative linguistic descriptions provides a more interpretable and effective signal for multimodal reasoning. By allowing the model to rely on qualitative assessments of trustworthiness, we can bridge the gap between abstract concepts and actionable insights in multimodal tasks.

\subsection{Has the Model Truly Recalled Visual Hint?}
\begin{figure}[t]
    \centering
    \includegraphics[width=0.48\textwidth]{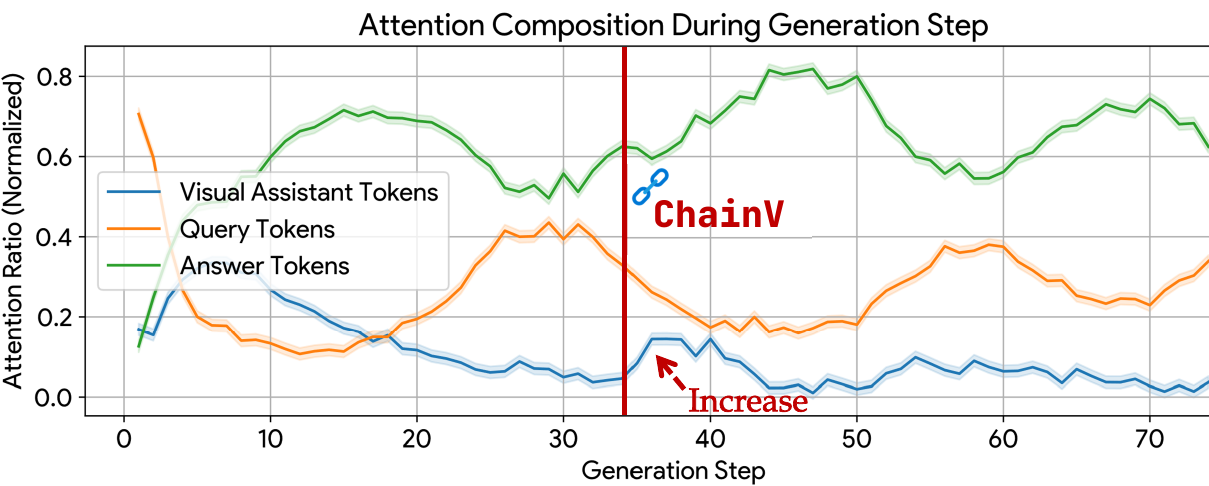}
    \vspace{-7mm}
    \caption{\textbf{Visualization of the received attention of visual assistant.} The reasoning case is randomly sampled from MIMO-VL-RL 7B on the MathVista benchmark. Best viewed in color.}
    \vspace{-6mm}
    \label{fig:vis}
\end{figure}

By utilizing attention signals obtained from the generated answers, our method effectively connects textual reasoning with visual evidence, creating a more reliable and interpretable multimodal reasoning pipeline. 
To further verify that our method indeed encourages the model to recall visual assistant information during reasoning, we visualize the variation of the received attention to the visual assistant $\vx_o$ across the generation steps. We randomly sampled a case from MIMO-VL-RL on the MathVista benchmark, as illustrated by the blue curve in Figure \ref{fig:vis}. It can be observed that at step $34$, the attention to the visual assistant begins to rise again, \textit{precisely when ChainV is inserted} into thinking process, demonstrating the effectiveness of our approach. This visualization provides clearer insights into the cross-modal interaction dynamics, revealing how our visual cues are selectively reactivated to guide and refine the reasoning trajectory toward the correct solution.

\section{Conclusion}
In this paper, we propose a novel training-free framework dubbed ChainV that utilizes reliability-scored atomic visual hints to make multimodal reasoning better and faster. By leveraging the cross-attention derived from the generated thinking tokens, we first select the answer-relevant visual patches, then convert the selected visual patches into three types of atomic visual hints, which are more semantically related to multimodal reasoning. In addition, we also introduce a consistency-based evaluation score to assess the reliability of the atomic visual hints and measure how faithfully it supports the thinking process. Finally, the pixel-level coordinates and the score of the visual hint are inserted into the reasoning process with a Bernoulli stochastic process. Detailed experiments across various baselines and datasets demonstrate the effectiveness of the proposed method.

\newpage
{
    \small
    \bibliographystyle{ieeenat_fullname}
    \bibliography{main}
}

\addtocontents{toc}{\protect\setcounter{tocdepth}{2}}
\clearpage
\setcounter{page}{1}
\maketitlesupplementary

\setcounter{section}{0}
\renewcommand{\thesection}{\Alph{section}}
\tableofcontents

\section{Implementation Details}
\label{sec:details}
All inference experiments are conducted on $8 \times$ NVIDIA A100 80GB GPUs. We run the models with the lmms-eval\footnote{https://github.com/XiaomiMiMo/lmms-eval} inference framework with vLLM\footnote{https://github.com/vllm-project/vllm}. To ensure reproducibility, we fix all random seeds and maintain a consistent \textbf{gpu memory utilization} setting across experiments, preventing fluctuations in inference latency. Reported latency values are averaged over five independent trials. The main inference configuration is as follows:

\begin{lstlisting}[style=jsonstyle, caption={\textbf{Inference Configuration}.}]
{
  "device": "cuda",
  "gpus": 8,
  "gpu_type": "A100-80G",
  "inference_framework": "lmms-eval",
  "max_new_tokens": 32768,
  "temperature": 0.0,
  "num_beams": 1,
  "do_sample": false,
  "top_p": 1.0,
  "batch_size": 8,
  "random_seed": 0,
  "numpy_seed": 1234,
  "torch_seed": 1234,
  "fewshot_seed": 1234
}
\end{lstlisting}
For GPT-based evaluation, we adopt the GPT-4o (\textbf{2024-05-13}) version to ensure consistency and reproducibility. In our experiments, we additionally inserted the following connective text into the query: \textit{``The second image is the visual OCR, highlighted by a red rectangle.''} This sentence was appended at the end of the input query to provide the model with explicit clarification about the role of the second image, ensuring consistent grounding between the visual OCR signal and the main reasoning process.

\section{Benchmarks}
\label{sec:ben_details}

\subsection{Reasoning Benchmarks}
We conduct comprehensive experiments on six widely used multimodal reasoning benchmarks as follows:

\textbf{MathVista.}~\cite{lu2023mathvista} MathVista assesses mathematical reasoning grounded in visual contexts, such as diagrams and plots. It integrates $28$ prior datasets and adds new subsets targeting logical, functional, and scientific figure reasoning. The benchmark challenges models to combine precise visual understanding with compositional calculations and multi-step quantitative reasoning, featuring a \textit{mini} version that contains $1,000$ samples for lightweight evaluation.

\textbf{MathVision.} \cite{wang2024measuring} MathVision is a benchmark for multimodal mathematical reasoning that involves visual information, such as diagrams, plots, and geometric figures. It contains $3,040$ problems sourced from real mathematical competitions, covering $16$ disciplines with five difficulty levels. The dataset tests a model’s ability to integrate visual understanding with symbolic computation and logical inference, requiring multi-step reasoning for the correct solution.

\begin{table*}[t]
\setlength\tabcolsep{6.5mm}
  \renewcommand{\arraystretch}{1.25}
\small
\centering
\caption{\textbf{Runs on MathVista task with various seeds.} The reported results include both accuracy and inference time.}
\label{tab:runs}
\begin{tabular}{c|c|cc|cc|cc}
\toprule
\multirow{2}{*}{\textbf{Run}} & 
\multirow{2}{*}{\textbf{Seed}} & 
\multicolumn{2}{c|}{\textbf{ChainV}} & 
\multicolumn{2}{c|}{\textbf{Baseline}} & 
\multicolumn{2}{c}{\textbf{Difference}} \\
\cmidrule(lr){3-4} \cmidrule(lr){5-6} \cmidrule(lr){7-8}
& & Acc(\%) & Time(s) & Acc(\%) & Time(s) & $\Delta$Acc & $\Delta$Time \\
\midrule
\rowcolor{mygray}
1   & 1234   & 83.3 & 1.7 & 81.4 & 3.5 & +1.9 & -1.8 \\
2   & 2000   & 83.4 & 1.6 & 81.4 & 3.0 & +2.0 & -1.4 \\
3   & 3000   & 83.3 & 2.0 & 81.5 & 2.8 & +1.8 & -0.8 \\
4   & 4000   & 83.3 & 2.3 & 81.5 & 3.6 & +1.8 & -1.3 \\
5   & 5000   & 83.2 & 2.2 & 81.4 & 3.2 & +1.8 & -1.0 \\
\bottomrule
\end{tabular}
\end{table*}

\textbf{WeMath.} \cite{qiao2024we} WeMath evaluates large multimodal models in visual mathematical reasoning. It comprises about $6500$ problems covering $67$ knowledge concepts across $5$ granularity levels, combining visual elements such as diagrams, geometric figures, and charts. Each problem is decomposed into sub-problems to analyze reasoning behaviors. It introduces a four-dimensional diagnostic framework to assess knowledge acquisition and generalization abilities.

\textbf{MMMU Pro.} \cite{yue2024mmmu} It aims to evaluate the ability to integrate and reason over both visual and textual information. The dataset comprises $10,500$ test questions, $900$ validation questions, and $150$ development questions, covering multiple disciplines. In this work, we adopt \textit{the vision version with 10-option multiple-choice questions}, which focuses on visual understanding and reasoning while expanding the number of candidate answers from four to ten.

\textbf{LogicVista.} \cite{xiao2024logicvista} It assesses the logical reasoning abilities of large multimodal models in visual contexts. It contains $448$ multiple-choice questions spanning $5$ reasoning tasks and $9$ underlying capabilities, with human-written reasoning annotations for each question. The problems involve visual elements, such as diagrams and puzzles, requiring the integration of perception and logical inference. LogicVista enables both multiple-choice and open-ended evaluations.

\textbf{OlympiadBench.} \cite{he2024olympiadbench} OlympiadBench is an Olympiad-level bilingual multimodal scientific benchmark comprising $8476$ problems drawn from high-level mathematics and physics competitions (including the Chinese college entrance exam) with both text and image modalities. Each problem is annotated with expert step-by-step solutions and metadata such as answer types and subfields, making it suitable for the evaluation of large multimodal models.

\subsection{General Benchmarks}
We conduct comprehensive experiments on six widely used multimodal understanding benchmarks as follows:

\textbf{VStar.} \cite{wu2024v}
VStar is a high-resolution multimodal benchmark designed to evaluate the \textit{visual search} and reasoning capabilities of MLLMs. It comprises a diverse collection of complex visual scenes paired with textual queries that require precise localization and fine-grained understanding. Each example is constructed to test the model’s ability to identify, attend to, and reason over critical visual regions rather than uniformly processing the entire image. 

\textbf{CV-Bench.} \cite{tong2024cambrian}
CV-Bench is a \textit{vision-centric} benchmark designed to evaluate the fine-grained visual reasoning capabilities of MLLMs. It consists of $2{,}638$ manually curated image–question pairs derived from standard vision datasets, covering both 2D and 3D understanding tasks. By emphasizing visual comprehension and cross-dimensional reasoning, CV-Bench provides a rigorous evaluation of vision-grounded reasoning and spatial awareness, complementing existing text-heavy multimodal benchmarks.

\textbf{ConBench.} \cite{zhang2024unveiling}
ConBench is designed to evaluate the consistency of MLLMs when presented with varying prompt solution spaces around the same knowledge point. It comprises approximately $1{,}000$ publicly sourced images and, for each image, constructs multiple discriminative question types (varying the solution space size) along with a generative question without a fixed ground truth. 

\textbf{ChartQA.} \cite{masry2022chartqa}
ChartQA is a large-scale multimodal benchmark for question answering over chart images, designed to challenge models with both visual understanding and logical reasoning. It contains $9.6$K human-written questions and $23.1$K machine-generated questions based on real-world bar, line, and pie charts. 

\textbf{SEED-Bench.} \cite{li2024seed}
SEED-Bench is a comprehensive benchmark for evaluating MLLMs across hierarchically defined capability levels. It comprises $\sim24$K multiple‐choice questions with high‐quality human annotations, spanning $27$ evaluation dimensions, including both comprehension and generation of text and images. 

\textbf{ScreenSpot.} \cite{jurmu2008screenspot}
ScreenSpot is a dataset originally introduced for resource discovery in smart spaces, comprising multidimensional GUI screenshots with annotated interactive elements. Each screenshot is paired with metadata about the target GUI element (\eg., icon, text, widget) and the spatial region related to user instruction or interaction cue. Over $1,200$ instructions from multiple platforms (iOS, Android, macOS, Windows, Web) have been annotated.

\begin{table*}[t]
    \small
    \centering
    \caption{\textbf{Reduction of Wait Tokens and Output Tokens on six multimodal reasoning benchmarks.}}
    \setlength\tabcolsep{3mm}
  \renewcommand{\arraystretch}{1.25}
    \begin{tabular}{l|cc|cc|cc|cc|cc|cc}
    \toprule[0.95pt]
    \multirow{2}{*}{\textbf{Method}} 
    & \multicolumn{2}{c|}{\textbf{MathVista}}
    & \multicolumn{2}{c|}{\textbf{MathVision}}
    & \multicolumn{2}{c|}{\textbf{WeMath}}
    & \multicolumn{2}{c|}{\textbf{MMMU Pro Vis}}
    & \multicolumn{2}{c|}{\textbf{LogicVista}}
    & \multicolumn{2}{c}{\textbf{Olympiad}} \\
    \cmidrule(lr){2-3} \cmidrule(lr){4-5} \cmidrule(lr){6-7} 
    \cmidrule(lr){8-9} \cmidrule(lr){10-11} \cmidrule(lr){12-13}
    & Output & Wait & Output & Wait & Output & Wait & Output & Wait & Output & Wait & Output & Wait \\
    \midrule
    \codexp & 778.7 & 9.9 & 4280.3 & 46.6 & 943.9 & 16.1 & 2646.4 & 23.2 & 1382.4 & 17.5 & 6432.7 & 9.1 \\
    \alphaexp  & 729.4 & 8.9 & 4387.7 & 43.9 & 1178.8 & 14.8 & 2802.3 & 23.0 & 1663.7 & 19.9 & 5624.5 & 8.0 \\
    
    \rowcolor{mygray}
    \chainvexp  & 710.2 & 6.5 & 3977.4 & 41.2 & 982.4  & 12.5  & 2313.7 & 20.3 & 1498.7 & 16.5 & 5164.8 & 7.3 \\
    \bottomrule[0.95pt]
    \end{tabular}
\end{table*}

\section{Statistical Significance Analysis}
\label{sec:ss}
To further verify the robustness of the performance improvement and efficiency gains observed in the main experiments, we conduct comprehensive significance testing across multiple random seeds on representative benchmarks. Using MathVista as a case study, we perform five independent runs with different random seeds (1234, 2000, 3000, 4000, 5000), with the same methodology applied to other benchmarks. The detailed results are presented in Table \ref{tab:runs}, and the statistical analysis reveals:
\begin{center}
(1) \textcolor{runpei-orange}{\textbf{Accuracy Analysis}}
\begin{itemize}
    \item \textbf{Mean Improvement}: +1.86\%
    \item \textbf{Standard Deviation}: $\pm$ 0.08\%
    \item $p$-\textbf{value}: $< 0.001$
\end{itemize}
\end{center}
\textbf{Accuracy Consistency}: ChainV demonstrates remarkable stability in accuracy improvement, achieving a mean gain of +1.86\% over the baseline with an exceptionally low standard deviation of $\pm$0.08\%. The consistency across all five runs (improvements ranging from +1.8\% to +2.0\%) underscores the method's reliability.
\begin{center}
(2) \textcolor{kaiming-green}{\textbf{Inference Time Analysis}}
\begin{itemize}
    \item \textbf{Mean Reduction}: -1.26s
    \item \textbf{Standard Deviation}: $\pm$ 0.36s
    \item $p$-\textbf{value}: $< 0.01$
\end{itemize}
\end{center}
\textbf{Efficiency Enhancement}: Beyond accuracy gains, ChainV significantly reduces inference time by an average of 1.26 seconds per sample (standard deviation: $\pm$0.36s), representing substantial improvements in computational efficiency.

In summary, both improvements are statistically significant ($p < 0.001$ for accuracy, $p < 0.01$ for time reduction), confirming that the observed advantages are not due to random variation.
This comprehensive analysis provides \textit{strong statistical evidence} that ChainV delivers consistent, reproducible, and statistically significant improvements across multiple runs, highlighting its robustness and reliability in enhancing vision-language model performance while maintaining computational efficiency.

\section{Analysis of Output Tokens Reduction}
The efficiency analysis of MiMo-VL-RL across six reasoning benchmarks reveals the significant advantages of ChainV in token optimization from multiple perspectives:

\textbf{Wait Token Reduction: Universal Efficiency Gains.}
ChainV demonstrates consistent and substantial reductions in wait tokens across all benchmarks, achieving the lowest wait counts in 5 out of 6 tasks. The most notable improvements include MathVista with 6.5 wait tokens (34.3\% reduction vs. \codexp, 27.0\% vs. \alphaexp), WeMath with 12.5 wait tokens (22.4\% reduction vs. \codexp, 15.5\% vs. \alphaexp), and MMMU Pro Vis with 20.3 wait tokens (12.5\% reduction vs. both baselines). This universal reduction validates that the visual cues in our method effectively enhance model confidence and optimize the reasoning flow.

\textbf{Output Token Optimization: Strategic Conciseness.}
By incorporating additional visual cues during the reasoning process, ChainV achieves a significant reduction in output length. The method produces the shortest outputs on both MathVista (710.2) and MathVision (3977.4) tasks. Notably, even without excluding the inserted reasoning statements (approximately $30+$ tokens), its actual response content remains the most concise across all benchmark evaluations. This balance between explanatory depth and response efficiency fully demonstrates its exceptional efficiency.

\section{Analysis of TLOPs Reduction}
To further evaluate the efficiency of ChainV, we compute the inference-time TFLOPs of MIMO-VL-RL on MathVista as an example. Although the \textit{dual-image setting} in ChainV introduces more visual tokens, the overall computation is significantly reduced during the autoregressive decoding stage, where the cost grows linearly with the accumulated sequence length. As a result, the original model requires about $19.08$ TFLOPs, while our more compact output setting only needs $16.09$ TFLOPs, achieving a reduction of $2.99$ TFLOPs ($\sim 15.7\%$) in total computation.

\section{Broader Impact}

\textbf{Accessibility and Practical Deployment.} ChainV significantly reduces computational costs in multimodal reasoning, enabling the deployment of advanced vision-language models across resource-constrained scenarios. This efficiency improvement facilitates broader application in educational, healthcare, and real-time interactive systems.

\textbf{Environmental Sustainability.} By substantially decreasing token generation during inference, ChainV directly lowers energy consumption and computational resource demands. This reduction supports the development of environmentally responsible AI systems and contributes to greener large-scale model deployment.

\textbf{Democratization and Transparency.} The training-free framework lowers barriers for implementing efficient reasoning, particularly benefiting researchers and organizations with limited resources. Furthermore, integrating visual grounding promotes reasoning interpretability, advancing the development of more transparent multimodal AI.

\section{Qualitative Case Studies}
\label{sec:more_case_studies}
This section presents qualitative case studies of ChainV, including both successful and unsuccessful cases across different models and reasoning benchmarks. The examples indicate that frequently appending our visual hint after \texttt{\textbackslash n\textbackslash n} can effectively support the ongoing reasoning step, thereby helping the model achieve better performance. Two representative examples that illustrate this phenomenon are provided in the following pages for detailed examination.

\begin{figure*}[t]
    \centering
    \includegraphics[width=0.89\textwidth]{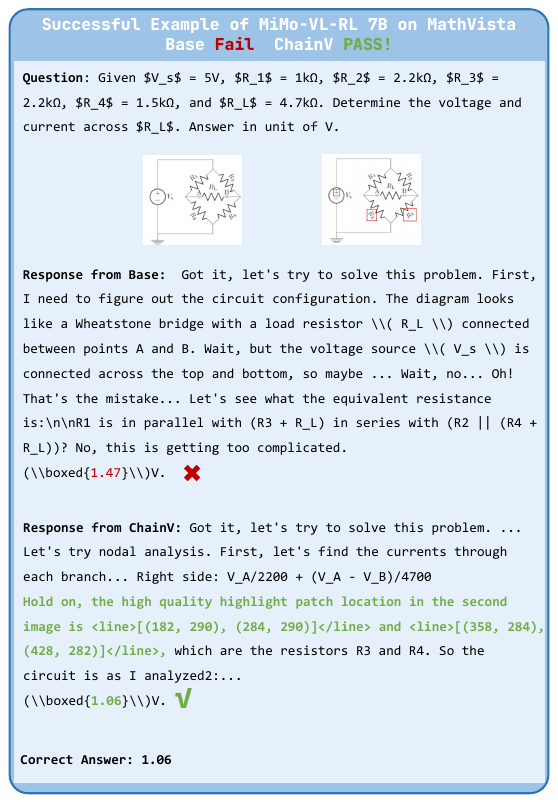}
\end{figure*}

\begin{figure*}[t]
    \centering
    \includegraphics[width=0.89\textwidth]{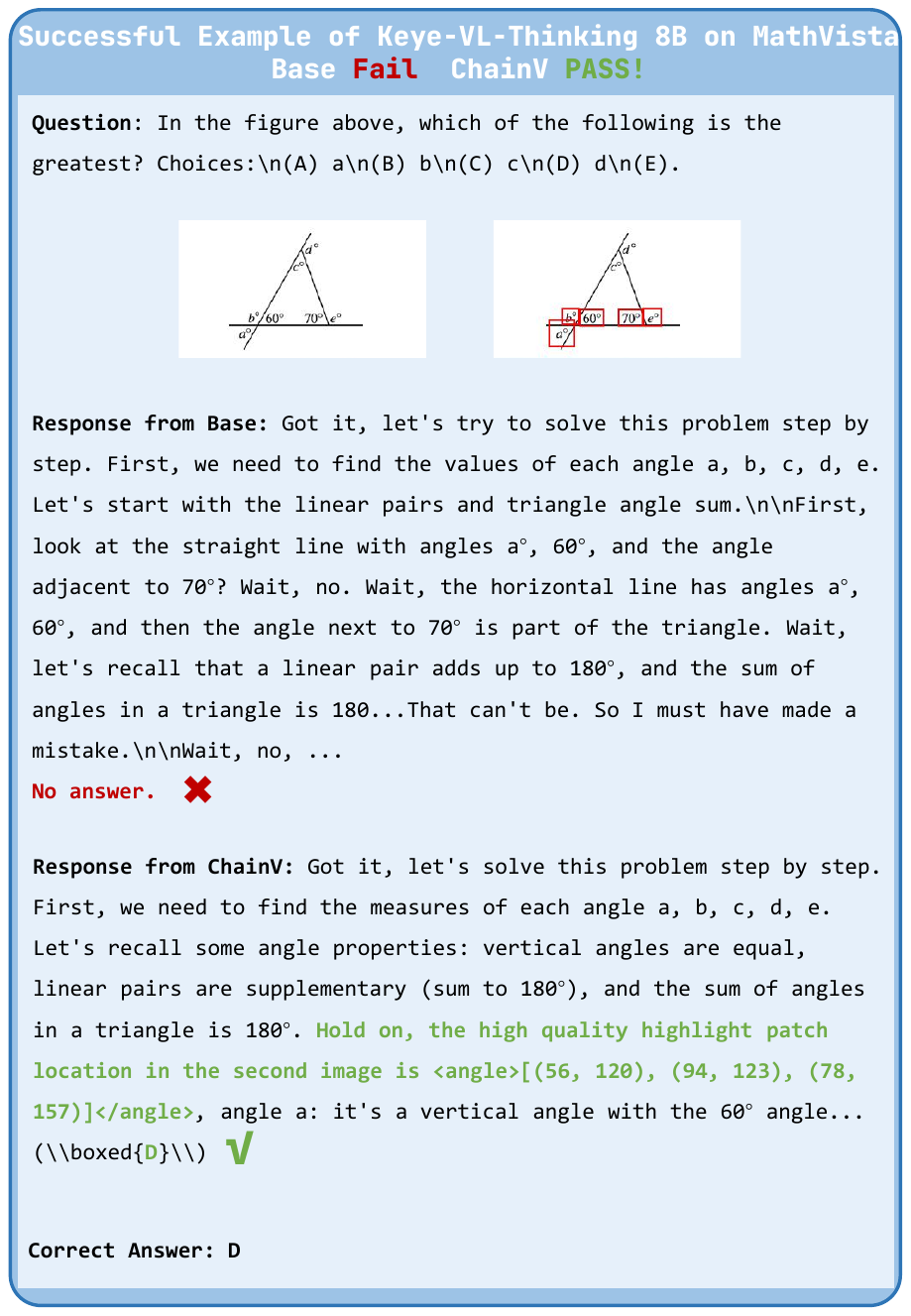}
\end{figure*}

\begin{figure*}[t]
    \centering
    \includegraphics[width=0.89\textwidth]{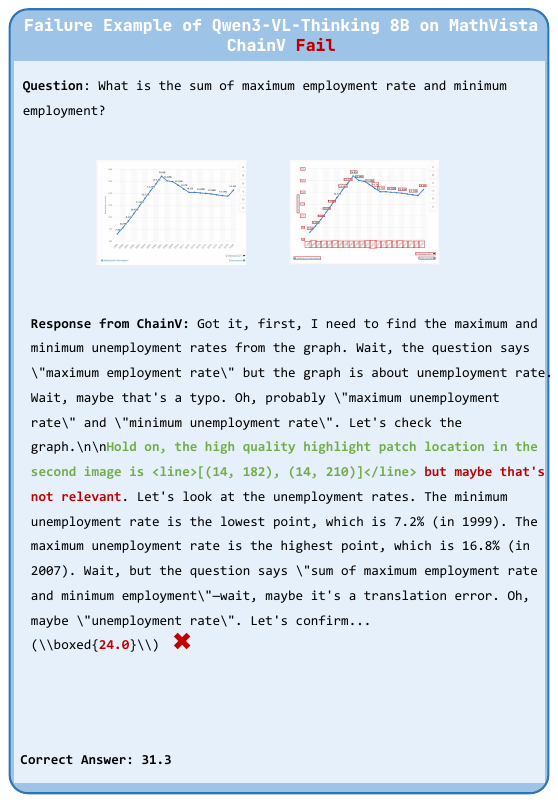}
\end{figure*}

\end{document}